\documentclass[conference]{IEEEtran}
\IEEEoverridecommandlockouts
\usepackage{cite}
\usepackage{amsmath,amssymb,amsfonts}
\usepackage{algorithmic}
\usepackage{graphicx}
\usepackage{textcomp}
\usepackage{xcolor}
\def\BibTeX{{\rm B\kern-.05em{\sc i\kern-.025em b}\kern-.08em
    T\kern-.1667em\lower.7ex\hbox{E}\kern-.125emX}}

\usepackage{xcolor}
\usepackage[algo2e,ruled,linesnumbered]{algorithm2e}
\usepackage{setspace}
\usepackage{enumitem}
\usepackage{verbatim}
\usepackage{amsmath}
\usepackage{amsfonts}
\usepackage{float}
\usepackage{array}
\usepackage{wrapfig}
\usepackage{framed}
\usepackage{blkarray}
\usepackage{graphicx}
\usepackage{dblfloatfix}

\newtheorem{theorem}{Theorem}[section]

\begin{document}

\title{How Powerful are Decoder-Only Transformer Neural Models?
}
\author{\IEEEauthorblockN{Jesse Roberts}
\IEEEauthorblockA{\textit{Department of Computer Science} \\
\textit{Vanderilt University}\\
Nasvhille, TN \\
Jesse.Roberts@Vanderbilt.edu}
}

\maketitle

\begin{abstract}

In this article we prove that the general transformer neural model undergirding modern large language models (LLMs) is Turing complete under reasonable assumptions. This is the first work to directly address the Turing completeness of the underlying technology employed in GPT-x as past work has focused on the more expressive, full auto-encoder transformer architecture. From this theoretical analysis, we show that the sparsity/compressibility of the word embedding is an important consideration for Turing completeness to hold. We also show that Transformers are a variant of B machines studied by Hao Wang.

\end{abstract}


\section{Introduction}

Transformer models have achieved state of the art performance on many tasks since their introduction in \cite{vaswani2017attention}. However, the provenance of their capabilities is not yet well understood. While some evidence suggests that capabilities may emerge as a function of model size \cite{wei2022chain}, continually increasing the number of parameters consumes significant energy posing risk to the environment \cite{rillig2023risks}. In this paper, we provide a proof that suggests that decoder-only transformer language models, like GPT-x, do not require the vast number of layers, attention heads, and parameters typical in current implementations to achieve powerful computation. 

The transformer architecture introduced in \cite{vaswani2017attention} is based on a denoising auto-encoder scheme. Interestingly, the work on these \textit{vanilla} transformers has largely been eclipsed by variations of the transformer like that in \cite{liu2018generating}, \cite{radford2018improving} (GPT), and \cite{devlin2018bert} (BERT). Much of this may be due to GPT-4 \cite{OpenAI2023GPT4TR} and its predecessors which have captured public attention. While the exact architecture of GPT-4 is closed source, GPT-3 and GPT-2 are known to be decoder-only transformer architectures \cite{radford2019language, brown2020language}. 

Work regarding the computational expressivity of the \textit{vanilla} transformer has proven it to be Turing complete \cite{perez2019turing,bhattamishra2020computational}. However, in \ref{sec:disambig} we show that these proofs do not naturally extend to the decoder-only transformer architecture. Further, no formal evaluation of the computational expressivity exists for the decoder-only transformer architecture. In this paper: 

\begin{enumerate}
    \item We show that the decoder-only transformer architecture is Turing complete
    \item We show that this result holds even for single layer, single attention head decoder-only architectures
    \item We establish a minimum vector dimensionality, relative to the token embedding size, necessary for Turing completeness
    \item We classify decoder-only transformer models as a causal variant of \textit{B machines} \cite{wang1957variant} 
    \item We provide an explanation for parameter inefficiency 
\end{enumerate}

\begin{figure}[b!] 
    \centering
    \includegraphics[width=\columnwidth]{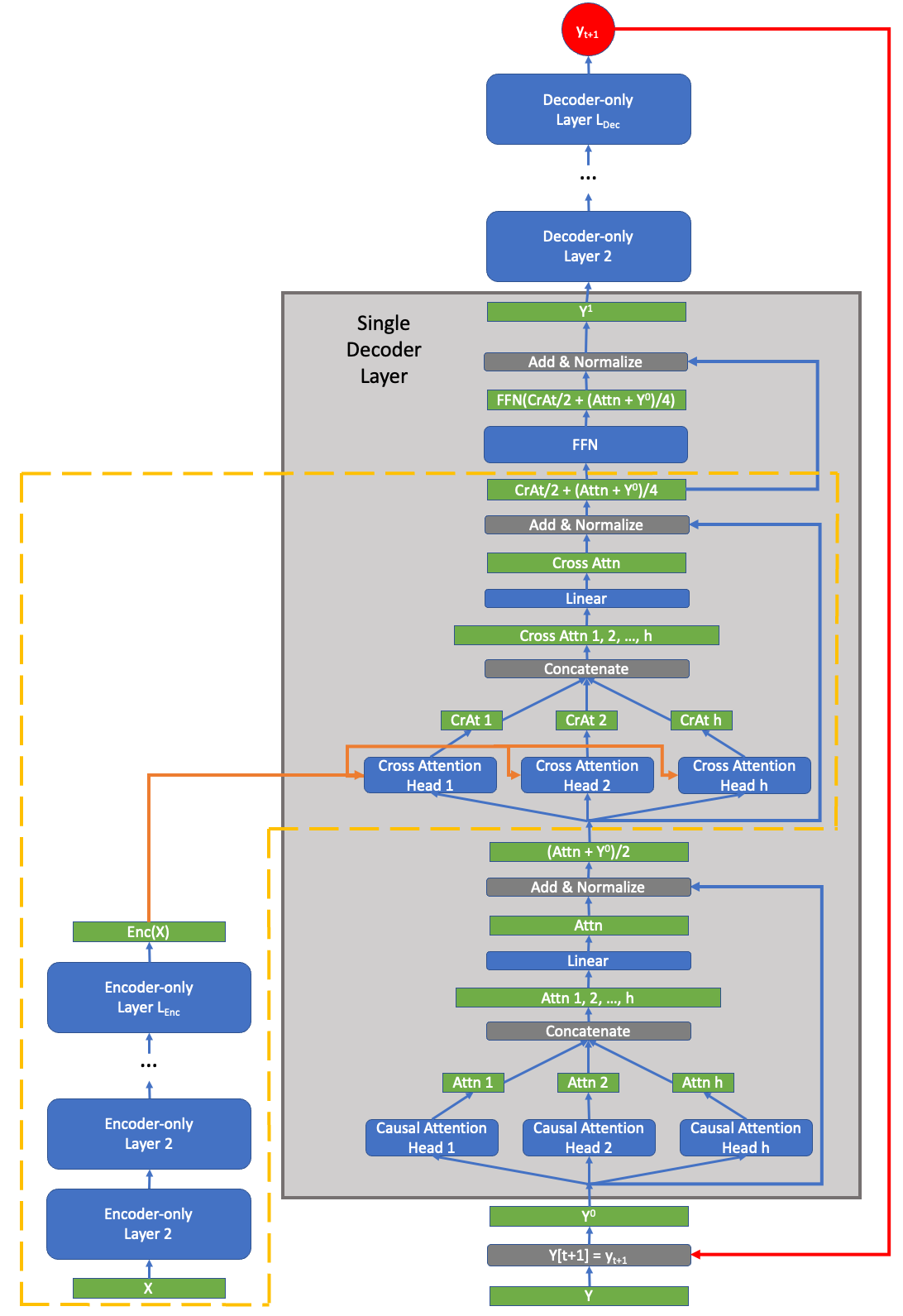}
        \caption{Vanilla Transformer Architecture. The yellow dashed line surrounds the sections removed to create a Decoder-only Transformer model. 
    \label{fig:Vanilla}}
    
\end{figure}

\begin{figure*}[h]
    \centering
    \includegraphics[width=0.7\textwidth]{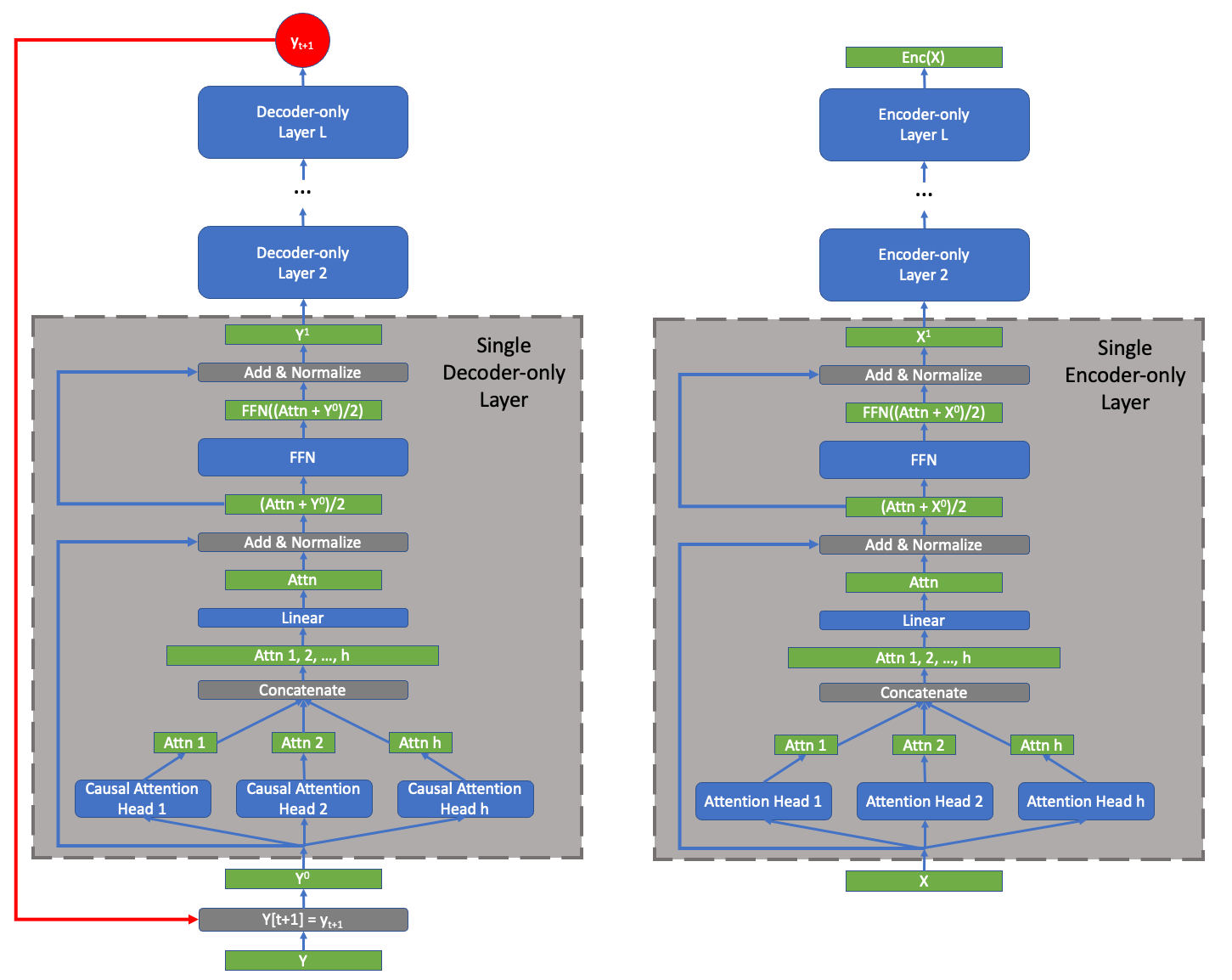}

    \caption{Decoder-only (left) and Encoder-only (right) Transformer Architectures. Green boxes are sequences of vectors with the width of the box representing relative sequence length. Red denotes a single vector. Gray and blue boxes denote simple and compound operations respectively.\label{fig:Dec_Enc_Only}}

\end{figure*}
Based on our results, we suggest that decoder-only architectures do not necessarily require the large number of parameters typically allocated to perform the necessary computations to support complex NLP functionality. Rather, the number of parameters may be necessitated by the interaction between the language modeling task and the architecture. This suggests that minor architectural adjustments could permit more parameter-efficient future models.

\section{Background}

\subsection{Disambiguating Decoder-Only Transformer Models}

Following after \cite{liu2018generating}, the creators of GPT refer to their architecture as a decoder-only transformer. Seemingly in contrast, the creators of BERT refer to it as an encoder-only model \cite{devlin2018bert}. This decoder-only/encoder-only architecture dichotomy is somewhat misleading as the two are architecturally identical as can be seen in \ref{fig:Dec_Enc_Only}. The differentiation lies in how the models execute. BERT and other encoder-only architectures are incapable of recursion. On the other hand, at each time step $t>0$, decoder-only architectures have access to their own outputs from all previous time steps. This permits the model to be trained to generate output auto-regressively. 

For brevity we follow previous conventions and refer to the transformer architecture presented in \cite{vaswani2017attention} as the \textit{vanilla} transformer, shown in \ref{fig:Vanilla}. Encoder-only transformer architectures do not possess a decoder. Similarly, decoder-only models do not have an encoder. These architectures are shown in \ref{fig:Dec_Enc_Only}. Notice, in the case of encoder-only models, disconnection at the encoder output is sufficient to unambiguously define the modification to the vanilla transformer architecture. This is not the case for decoder-only architectures. 


\subsubsection{Modifying the Vanilla Transformer to form a Decoder-only Model}

To create a decoder-only model, the vanilla architecture is modified in two ways. First, the connection to the encoder is removed. Second, the cross-attention which allows the decoder to conditionally attend to the encoder output at each layer of the decoder is eliminated. These, along with the entire encoder, are surrounded by a dashed yellow line in \ref{fig:Vanilla} to visualize what is eliminated. As mentioned previously, this superficially suggests that encoder-only and decoder-only architectures are identical as seen in \ref{fig:Dec_Enc_Only}.

\subsubsection{Differentiating Encoder-only and Decoder-only Models}

\label{sec:disambig}

Decoder-only models have three necessary characteristics which are derived from their function in the vanilla transformer. The decoder must (1) provide a means of auto-regressively predicting the next token based on the tokens generated so far given the encoder input as contextualization. In \ref{fig:Dec_Enc_Only} this is shown as the recursive red connection mapping the output vector back into the last element of the input sequence of vectors. To be suited to this task, decoder-only models must (2) not see future values when evaluating a query on the input sequence of vectors. This is why decoder-only models are often referred to as causal language models (CLM). In \ref{fig:Dec_Enc_Only}, we refer to the decoder attention heads as causal attention heads rather than masked attention heads as they are called in \cite{vaswani2017attention}. The model must be (3) trained to predict the next token given the current input sequence of vectors. This training method coupled with recursion allows decoder-only models to auto-regressively generate arbitrarily long (up to the max size of the input vector sequence) sequences. 

If any of the above are violated, the model can't be reasonably considered a decoder-only model as it is no longer capable of auto-regressive next token prediction.

\subsection{Related Theoretical Work on Transformers}

Transformers were shown to be Turing complete first in \cite{perez2019turing} with a simpler approach to the proof given in \cite{bhattamishra2020computational}. The latter is based solely on the ability of the transformer to simulate arbitrary RNNs which are known to be Turing complete \cite{siegelmann1992computational}. This latter work also considered the contribution of the various architectural elements to the computational power. In their construction, they find the computational universality of the transformer is maintained even if the encoder acts essentially as an identity operator for the appropriate input. All significant computation, beyond input presentation, is handled exclusively in the decoder and FFN. However, in both proofs, the encoder is a necessary component without which the Turing completeness result does not hold. 

Hahn shows that softmax based attention is often well approximated by the hardmax function \cite{hahn2020theoretical}. They further show that one can apply input restrictions to transformers such that PARITY is unrecognizable in a single feedforward encoder pass regardless of the number of layers. However, their work assumes the number of computations is bounded by the length of a bounded length input. 

In \cite{yun2019transformers}, the authors studied encoder-only architectures and showed that they were capable of universal function approximation. For this to be the case, the attention mechanism of the encoder-only architecture must be sufficient to provide the FFN with access to all subsets of the input field. Or to put this in terms familiar to a convolutional system, the attention mechanism must be capable of implementing any arbitrary feature map. This result is also important to the theoretical understanding of decoder-only transformer architectures as is clear in \ref{fig:Dec_Enc_Only}. Specifically, this implies that decoder-only models are universal function approximators for the $n$\textsuperscript{th} attention query in the $L$\textsuperscript{th} layer given an input sequence of length $n$. However, this does not prove Turing completeness.

It is reasonable to believe universal function approximation may be grounds for expecting Turing completeness to hold due to the progression of the literature for ANNs which began by showing universal function approximation \cite{hornik1989multilayer} and then progressed, through the addition of recursion, to Turing completeness \cite{siegelmann1992computational}. Further, it is intuitive based on the recursive capability of decoder-only models coupled with universal function approximation, as a model which can compute any partial recursive function is necessarily Turing complete \cite{turing1937computability}. However, this would require that the computational class which includes primitive functions with composition and minimisation \cite{neto1997turing} be equivalent to the class of universal function approximators. Interestingly, the equivalency of these classes has never been addressed, leaving this an open question.

The only research regarding the computational expressivity of decoder-only transformer models (at the time of writing) is that of \cite{schuurmans2023memory}. They recently considered the computational power of memory augmented language models. They showed that, when augmented by a memory module which is not part of the typical decoder-only transformer architecture, the model is Turing complete. To date, no work in the literature has addressed the computational power of typical decoder-only transformer models.

\subsection{Required Conventions Inherited from Vanilla Transformers}

\label{sec:Requirements}

The following are not architectural or training limitations, and are instead conventions that could be relaxed by future transformer architectures. However, we choose to evaluate the computational expressiveness of the typical decoder-only transformer model in common use.  

First, the input embedding and output embedding used in the decoder must be identical. This permits the model output to be directly appended to the input vector sequence. Implicitly, this means decoder-only models can't have orthogonal input and output dimensions in the context vector. This is an important point as the proof method from \cite{bhattamishra2020computational} requires orthogonality which was permitted by cross-attention. However, cross-attention is removed in the decoder-only model as seen in \ref{fig:Vanilla}.  

Second, the input dimension of the FFN(s) must have the same dimensionality as the model dimension ie. the dimensionality of a vector in the input sequence. This disallows \textit{sparsification} in the latent space which could be used to create a FFN input dimensionality greater than the model dimensionality. However, this does not require that the dimensionality of the model, $d_{model}$,  be equal to the embedding dimensionality, $d_{embed}$.

\section{Definitions \& Approach}

We modify the formalism established in \cite{perez2019turing} and used in \cite{bhattamishra2020computational} for theoretical transformer analysis to be appropriate for our analysis of decoder-only architectures.

\subsection{Embedding \& Position}

Transformers embed inputs as higher dimensional vectors via a base embedding $f_b$. So, for a vocabulary $\Sigma$ with cardinality $m$, $f_b:\Sigma \rightarrow \mathbb{Q}^{d_b}$ where $d_b$ is the number of dimensions in the embedding. 

The Turing complete proof method will require that the transformer recognize the RNN stop token. Therefore, we define the embedding for the end symbol $\$$ such that $f_b( \$ ) = \mathbf{1}^{d_b}$. 

In most transformer architectures the embedding is supplemented with positional information (whether explicitly defined or learned). Here we define the positional encoding as $pos:\mathbb{N} \rightarrow \mathbb{Q}$. So, for a vector $\mathbf{S}_k = (\sigma_1, . . . , \sigma_k)$ with $\sigma_k \in \Sigma$ for all $k \geq 1$, the embedding with position of $\mathbf{S}_k$ is given by $(f_b(\sigma_1)+\nolinebreak pos(1),...,f_b(\sigma_k)+\nolinebreak pos(k))$. The dimensionality of the combined token and position embedding is $d_{embed}$.

\subsection{Decoder-only Transformer Architecture}

A single layer decoder-only transformer is comprised of multi-headed attention followed by a feed forward network as seen in \ref{fig:Dec_Enc_Only}. It takes as input a sequence $\mathbf{Y} = (\mathbf{y}_1, . . . , \mathbf{y}_k)$ of vectors where $k \geq 1$. The output of any single layer is likewise a sequence of vectors $\mathbf{Z} = (\mathbf{z}_1, . . . , \mathbf{z}_k)$. 

As previously mentioned, all $\mathbf{y} \in \mathbf{Y}$ and $\mathbf{z} \in \mathbf{Z}$ must have dimensionality $d_{model}$. However, $d_{embed}$ is not required to be equal to $d_{model}$. We choose to include additional space in $d_{model}$ such that the overall representation is sparse. Specifically, $d_{model} = 2 \cdot d_{embed} + 3$. The details of this choice are discussed in the proof.

The full decoder-only transformer architecture is formed by a stack of $L$ layers, each composed of a single layer decoder. The output of a single execution of the model is a single vector $\mathbf{z}^L_k$, where superscript $L$ denotes the $L^{th}$ layer. This vector is then directly appended to $\mathbf{Y}$ such that $\mathbf{y}_{k+1} = \mathbf{z}^L_k$. The output of the previous execution is appended to the input of the subsequent execution, creating recursion. 

The model will recursively execute continuously until a stopping criteria is met. Typically, the model is allowed to execute until a special token embedding is output by the model. After execution terminates, the size of the output sequence will be $|\mathbf{Y}| = k + N$ where $N$ is the number of executions. The sub-vector $(\mathbf{y}_{k+1},..., \mathbf{y}_{k+N})$, referred to as the \textit{response}, is the complete output of the model given the original \textit{prompt} contained in $(\mathbf{y}_{1},..., \mathbf{y}_{k})$.

\subsection{Self-Attention}

Every layer in \ref{fig:Dec_Enc_Only} has one or more causal, self-attention, heads which filter the prompt to attend to the germane portions. Each attention head possesses functions $Q(\cdot)$, $K(\cdot)$, and $V(\cdot)$ which apply a linear transformation to each $\mathbf{y} \in \mathbf{Y}$. This results in a sequence of query vectors $\mathbf{Q}$, sequence of key vectors $\mathbf{K}$, and sequence of value vectors $\mathbf{V}$. 

Each head creates a filtered view of the layer input given each query. Value vectors in $\mathbf{V}$ are chosen using the query vector $\mathbf{q}$, the sequence of keys $\mathbf{K}$, and scoring function $f^{att}(\mathbf{q},\mathbf{k})$ $\forall \mathbf{k} \in \mathbf{K}$. The scoring function is the dot product of the vectors combined with a non-linear function \cite{vaswani2017attention}. 

Specifically, $\mathbf{q}$ attends to $\mathbf{V}$ according to an attention vector $\mathbf{a} = hardmax(\alpha_1,...,\alpha_n)$ with $\alpha_i = f^{att}(\mathbf{q},\mathbf{k}_i)$ for all $1 \leq i \leq n$. Then, the $\mathbf{q}$ attention on $\mathbf{V}$ is $\langle \mathbf{a}, \mathbf{V} \rangle$. This self-attention is compactly referred to as $Att(\mathbf{q},\mathbf{K},\mathbf{V})$.

In \cite{vaswani2017attention}, softmax is used. However, hardmax is used in our case to ensure all outputs are rational. Specifically, for a vector $\mathbf{x}$ with $m$ maximum values, $hardmax(\mathbf{x_i})= 1/m$ $\forall x_i \in \mathbf{x}$ iff $x_i$ is a maximum, else $hardmax(\mathbf{x_i})= 0$. 

In the case of multiple attention heads, each of these filtered views are concatenated and then agglomerated. Agglomeration is necessary because the concatenation step may produce a representation which no longer has dimensionality $d_{model}$. To return to $d_{model}$, a linear transformation using a set of weights,  $W^l$,  with dimensionality $d^v_{l,H}$x$d$ is applied. The concatenation and linear transform are referred to compactly as $\mathit{Conn}(\cdot)$.

\subsection{Feed Forward Network}

The feedforward network, referred to as $O(\cdot)$, is fully connected and parameterized by $\theta$. The output is $\mathbf{Z} = (\mathbf{z}_1, . . . , \mathbf{z}_k)$.

\subsection{Single Layer Decoder-Only Models}

The following set of equations fully characterizes the function of a single layer decoder-only transformer model. Notice that the output is a sequence of vectors.

\begin{align}
\mathbf{p} &= \mathit{Att}(Q(\mathbf{y}), K(\mathbf{Y}), V(\mathbf{Y})) \\
\mathbf{r} &= \mathit{Conn}(\mathbf{p}) + \mathbf{y} \label{eq:withresid} \\
\mathbf{z} &= O(\mathbf{r}:\theta) + \mathbf{r} \label{eq:singleOutput}
\end{align}

The set of equations characterizing a single layer are compactly referred to as $\mathit{Dec}_l(Y_l;\theta_l)$, with $l$ denoting the layer.

\subsection{Multi-Layer Decoder-Only Models}

A multi-layer decoder-only transformer has one or more additional layers which take the output sequence generated by the previous layer as as input. 

The output sequence of vectors from layer $l$ is then referred to as $Y^{l}$ and becomes the input to layer $l+1$. The output equation becomes $Y^{l+1} = \mathit{Dec}_l(Y_l;\theta_l)$, with $\mathbf{Y}_0 = \mathbf{Y}$. The output of a model is a single vector, the $k^{th}$ element of the output vector for the last layer.  

\subsection{Proof Approach}

Our approach to proving Turing completeness, following the example of \cite{bhattamishra2020computational}, is to show that a decoder-only transformer architecture is capable of simulating the computations performed by an RNN. Based on the work of \cite{siegelmann1992computational}, RNNs are known to be at least as computationally expressive as Turing machines. Therefore, if a decoder-only transformer model can simulate an arbitrary RNN, then the decoder-only transformer architecture is at least as computationally expressive as a Turing machine. 

Just as in \cite{bhattamishra2020computational} we will say that an RNN is simulated if (1) at each time step the input vector to the neural network contains the input $x_t$, (2) at each time step the input vector to the neural network contains the hidden state $h_t$, and (3) the decoder-only model stops at the same time step as the RNN. 

To simulate an RNN via a decoder-only transformer architecture we use the decoder to implement recursion as has been done previously for vanilla transformers. However, our construction is different in that decoder-only transformers do not have an encoder. Therefore, we will provide the input to the model as the \textit{prompt} and the \textit{response} will be appended until execution terminates. It is clear that $\mathbf{Y}$ will always contain $h_t$ and $x_t$ for all timesteps. We will show by construction that self-attention, a feedforward neural network, and recursion via the decoder-only transformer is sufficient to attend to and present $h_t$ and $x_t$ to the FFN for all $t$ and simulate an arbitrary RNN. 

\section{RNN Simulation by Decoder-Only Transformer}

In this section we prove that there exists a single-layer, single-attention head, decoder-only transformer which may simulate any RNN. Some details are encapsulated in theorems below the proof body. In the subsequent sections we give a detailed, intuitive explanation of the proof and discussion of the implications and limitations. 

\subsection{Proof}

Consider a decoder-only transformer with a single layer and single attention head in that layer. 

Before the first execution of the network, the input sequence of vectors, $\mathbf{Y}$, contains the \textit{prompt} (inputs to the RNN) in the form $\mathbf{y}_i = [f_b(\sigma_i),0^{d_{embed}},i,t,\mathit{stop}]$ with each $\mathbf{y}_i \in \mathbf{Y}$ having dimensionality $d_{model}$. The value of $i$, $t$, and $\mathit{stop}$ for $i\leq k$ are $\mathit{pos}$, 0, and 0 respectively. The penultimate element in the \textit{prompt}, $\mathbf{y}_{k-1}$, has $\sigma_{k-1}=\$$, and the last element, $\mathbf{y}_{k}$, has $\sigma_{k}=0^{d_{embed}}$, the RNN start token. 

Appropriate $Q$, $K$, and $V$ linear transforms are applied to each element of $\mathbf{Y}$ such that $\mathbf{q}_i = \mathbf{y}_i$, $\mathbf{k}_i = [0^{d_{embed}},0^{d_{embed}},1,-1,0]$, and $\mathbf{v}_i = \mathbf{y}_i$. Therefore,  $\langle \mathbf{q}_i, \mathbf{k} \rangle = i-t$. The existence of such a $Q$, $K$, and $V$ is trivial. 

By application of the nonlinear function, $f^{att}(\mathbf{q},\mathbf{k}_i)$, the attention on each $v \in \mathbf{V}$ is $\alpha_k = -|i-t|$. Therefore, $\mathit{hardmax}(\mathbf{V}) = 1$ when $i=t$ and 0 $\forall i \neq t$. Therefore, $\mathit{Attn}(\mathbf{q}_{i=t},\mathbf{K},\mathbf{V}) = \mathbf{x}_{i=t}$. Therefore, the $t$ element in the query vector selects the $i=t^{th}$ element from the \textit{prompt}. 

To generate the $t^{th}$ element of \textit{prompt}, the query $\mathbf{q}_{k+t} = Q(\mathbf{y}_{k+t})$ is used. The model will execute a total of $N$ times such that $t=0...N$. 

Notice the first execution has $\mathbf{q}_{k+t} = [0^{d_{embed}},0^{d_{embed}},i=pos(k),t=0,\mathit{stop}=0]$ and, by application of the agglomeration and residual connection as described in \ref{lem:compressionX}, the vector presented to the FFN will be $[h_t = f_b(\sigma_{k}), x_t = f_b(\sigma_{t}),i,t,\mathit{stop}]$. The FFN will output the vector $\mathbf{y}_{k+t+1} = [h_{t+1}, 0^{d_{embed}},i=k+t+1,t=t+1,\mathit{stop}]$ which is appended to the sequence $\mathbf{Y}$. Therefore, for all executions $t>0$, the vector presented to the network will be $[h_t = f_b(\sigma_{i=k+t}), x_t = f_b(\sigma_{i=t}),i,t,\mathit{stop}]$. 

As proved in \ref{lem:recStop} and \ref{lem:override} there exists an FFN such that once the stop token, $f_b(\$)$, has been encountered the output of the FFN for all subsequent time steps will be $\mathit{stop}=1$ and the value $x_t = f_b(\sigma_{i=t})$ will be overridden in latent space such that for all $t>k$, $x_t = x_k = f_b(\$)$ due to \ref{lem:CombinableNeuralNetworks}. 

At all time steps the FFN will be presented with $x_t$, $h_t$, and will terminate based solely upon the weights of the RNN. Therefore, there exists a decoder-only transformer which may simulate any RNN.

\subsection{Theorems}

\begin{theorem}[Single Network replacement of Cascaded Networks]
\label{lem:CombinableNeuralNetworks}

For any pair of fully connected feed forward neural networks (FFNs) such that the outputs of the first are fed into the inputs of the next, there exists a single FFN whose outputs will be identical to the outputs of the second network. 

\end{theorem}

By construction, the output weights can be directed into the input of the subsequent network and stored in a single set of network connection matrices such that a single network is created. The outputs of the first network in the cascade become a latent space within the combined network.

\begin{theorem}[FFN Override Input]
\label{lem:override}

Given any neural network with inputs $x_1,...,x_{k}$, outputs $O={o_1,...,o_{k}}$, and $n_l$ neurons in $l$ hidden layers. We may add an input $x_{k+1}$ and neuron $n_l+1$ to hidden layers $1...l$ such that an arbitrary subset $o' \in O$ are overridden by the added neurons. 

\end{theorem}

All weights from input $x_{k+1}$ to neurons $1,...,n_1$ are set to zero. All weights from inputs $x_1,...,x_{k}$ to neuron $n_1+1$ are zero. The weight from input $x_{k+1}$ to neuron $n_1+1$ are set to infinity. 
 
In each layer $l>1$, neuron $n_l+1$ has connections set to zero for all neurons $1,...,n_{l-1}$. And in each layer $l>1$, $n_l+1$ has connections set to infinity for neuron $n_{l-1}+1$. This forms a column of mutually connected neurons.  
 
An arbitrary subset of outputs $o' \in O$ may be chosen which are to be affected by the added column of neurons. The weights connecting neuron $n_l+1$ in hidden layer $l$ to each output in $o'$ are set to infinity and the weights connecting $n_l+1$ to each output in $O\setminus o'$ are set to zero. 

For all neurons $n_l+1$ in all layers, the bias value is set to zero. Therefore, when the input $x_{k+1}=0$, the original function of the network is left unchanged. When $x_{k+1}=1$, the value of each output in $o'$ is forced to be the max activation function value.

\begin{theorem}[Recognize the stop token]
\label{lem:recStop}

Given a stop token $\$$ that is embedded as a vector with $k$ elements each equal to $1$ and presented to a neural network along inputs $x_1,...,x_{k}$, a neuron may be defined such that the output is non-zero only for inputs that are $\epsilon$ close to the stop token embedding. 

\end{theorem}

Since the stop token is defined as a vector of ones, for any token presented, the output of the neuron is zero when $k-\Sigma_{i=1}^k x_i>\epsilon$ and is greater than zero for all other inputs so long as the bias is $b=\epsilon - k$. By setting the output weight of the neuron to be a large value, any non-zero output will result in saturation of downstream neurons with non-zero connecting weights. Therefore, an output that represents whether the stop token has been presented will have a max activation function value iff the input along $x_1,...,x_{k}$ is within $\epsilon$ of $\$$.

\begin{theorem}[Compression of $\mathbf{x}_t$ and $\mathbf{h}_t$ into $\mathbf{r}_t$]

\label{lem:compressionX}

Given a token base embedding with dimensionality $d_{embed}$, $\mathbf{x}_t$ and $\mathbf{h}_t$ may be losslessly compressed into $\mathbf{r}_t$ when each have dimensionality $d$.

\end{theorem}

Recall the dimensionality of $V(\mathbf{x}_t)$ with $\mathbf{x}_t \in \mathbf{Y}$ is not related to $d$. Rather, $V:\mathbb{Q}^d \rightarrow \mathbb{Q}^{d_{embed}}$ such that $V(\mathbf{x}_t) = \mathbf{\sigma}_t$ with $\mathbf{\sigma}_t$ being a token in $\Sigma$. Then, by matrix multiplication with $W$ defined as:
\vskip-0.5em
\scriptsize
\[
\begin{blockarray}{cccccccccc}
 &  & (1,d_{embed}) &  &  &  & & &  (1,d_{model})& \\
\begin{block}{(ccccccccc)c}
0       &    ...  &  0    & 1   & 0      & ...& 0  & 0  & 0      & \\
\vdots  & \ddots  & 0     & 0   & \ddots & 0  & 0  & 0  & \vdots & \\
 0      & 0       & 0     & ... & 0      & 1  & 0  & 0  & 0      &  \\
\end{block}
& & & & & & & & (d_{embed},d_{model}) & 
\end{blockarray}
 \]
 
\vskip-2em
 \normalsize

 The resulting vector is $\mathit{Conn}(\mathbf{p}_{t}) = [0^{d_{embed}}, \mathbf{x}_t, 0, 0, 0]$. Finally, by applying the residual connection we have $\mathbf{r}_t = [\mathbf{h}_t, \mathbf{x}_t, i, t, \mathit{stop}]$.

\section{Proof Explanation}

To accomplish RNN simulation, an attention head is used to select the appropriate input from the \textit{prompt} in $\mathbf{Y}$. The attention head and agglomeration weights shift the embedded representation of the input into an empty area of the model embedding. Then, the residual connection sums the input vector with the attention representation. This results in $h_t$ and $x_t$ in a single vector of size of $d_{model}$. This vector is then presented to the FFN which contains the RNN weights as well as cascaded supplementary functions.

\subsection{Vector Elements}

Recall the base embedding has dimensionality $d_{embed}$. As discussed previously, the input dimension of the FFN must be $2 \cdot d_{embed} + 3$. From the requirements inherited from transformer conventions, the model dimension must be equivalent to the input dimension of the FFN. So, we choose $d_{model} = 2 \cdot d_{embed} + 3$. Therefore, each $\mathbf{y} \in \mathbf{Y}$ is composed as $\mathbf{y}_i = [f_b(\sigma_i),0^{d_{embed}},i=pos,t,\mathit{stop}]$. $f_b(\sigma_i)$ is the base embedding of the token in position $i^{th}$ position. $0^{d_{embed}}$ is the unused space to permit simulataneous presentation of $x_t$ and $h_t$ to the FFN. The sequence position of $\sigma_i$ is stored in $i$ and the execution time step is written by the FFN to $t$.

\subsection{Attention}

A single atttention head attends to $\mathbf{y}_i$ where $i=t$. This input value is referred to as $x_t$ as this is the value which would be presented to an RNN at time $t$. 

The attention head will return $x_t$ with size $d_{embed}$. By application of a linear transformation, $W^l$, $x_t$ is right shifted $|d_{embed}|$ elements and padded with zeros to have dimension $d_{model}$. Finally, via the residual connection and normalization, the resulting $\mathbf{r}_t$ from \ref{eq:withresid} is $\mathbf{r}_t = [h_t, x_t, i, t, \mathit{stop}]$, proved in \ref{lem:compressionX}. 

Note that for all $t > k$, the attention head will select a value from the \textit{response}. If unaddressed, this would prevent RNN simulation as only the \textit{prompt} contains RNN input. However, as explained, when $t>k$ the stop token will have been seen and the FFN will ignore the value presented by the attention head by overriding it with the stop token. As an alternative construction, the position encoding could be set to zero for all vectors in the \textit{response} generated by the model as this would result in the stop token being attended to for all $t>k$. However, we avoid this solution as it is a significant deviation from typical models.

A similar method for selection of the $t^{th}$ element of $\mathbf{y}$ is used in  \cite{bhattamishra2020computational}. However, in their construction the attention head is performing cross attention rather than causal, self attention \ref{fig:Vanilla}. This important difference means that their construction does not apply to decoder-only transformer models.

\subsection{FFN Operations}

The FFN instantiates the weights of the RNN. However, the FFN has three additional functions. The FFN (1) recognizes the RNN stop token and (2) overrides the $x_t$ provided by the attention head with the stop token if the $\mathit{stop}=1$. Lastly, the FFN (3) acts as a counter which generates the execution timestep, $t+1$, based on $t$ in the previous input vector.

The stop token recognition, override function, and RNN weight instantiation are each proved possible for standalone networks. However, by considering each of the individual networks as cascaded, there exists a single network which may implement these three functions in series.

\subsection{Summary}

At each time step, the transformer FFN is presented with $x_t$ and $h_t$. Further, $h_{t+1}$ will generate the RNN stop token at the same time step as the RNN. This is because the RNN weights are a proper subset of the FFN weights and they have identical access to $x_t$ and $h_t$ as would occur in an RNN. There necessarily exists a decoder-only transformer capable of simulating an arbitrary RNN and thus the class of decoder-only transformer models is shown to be at least as computationally expressive as RNN models. Therefore, there exists a computationally universal decoder-only transformer.

\subsection{Assumptions \& Limitations}

There are 2 main assumptions required by this proof which limit applicability to general decoder-only models. 

First, the attention mechanism here uses hardmax as opposed to the typically used softmax. This assumption is similar to prior work in theoretical transformer analysis \cite{perez2019turing,bhattamishra2020computational} and is necessary to ensure values are kept rational which is not the case for softmax. Additionally, \cite{hahn2020theoretical} suggests that transformer softmax attention heads may focus attention on high scoring context and learn behavior that is well approximated by hard attention. 

Second, this work inherits the assumptions made in the proof of RNN Turing completeness. For the proof of RNN computational universality in \cite{siegelmann1992computational} to hold, infinite precision, infinite output space, and value rationality are required. 

These assumptions are typical in theoretical work regarding the transformer architecture. However, future work should seek to characterize transformer computational expressivity under relaxed assumptions.

\section{Discussion}

\subsection{Relationship Between Model Dimensionality and Turing Completeness}

Recall the requirements discussed in \ref{sec:Requirements}. An interesting consequence of these requirements is that, for a decoder-only transformer to be Turing complete, it must have dead space in the model dimension. That is, it must satisfy $d_{model} > d_{embed}$. This dead space is necessary to present both the last output, $h_t$, and the current input, $x_t$, to the FFN for computation of the next value in the sequence. Presentation of both values can't be guaranteed without satisfying the above inequality.

As brief proof by contradiction, assume that we are guaranteed to be able to present both $h_t$ and the $x_t$ without dead space in the model dimensionality. Since there is no dead space, every element in the vector is used to embed some piece of information about the token. To present both the embedding for the input and the last output to the FFN (without violating the mentioned requirements) we must compress the input or the last output. In the case of a dense embedding ie. a single bit, compression is not possible. Therefore, without the presence of dead space in the model dimensionality it may be impossible to present $x_t$ and $h_t$ to the FFN at time step $t$. Therefore, assuming no dead space is required to present both $h_t$ and $x_t$ at a single time step leads to a contradiction.

In the case of simulating an RNN, we can say that the minimum model dimension for $x_t$ and $h_t$ to be presented to an RNN simulating FFN simultaneously must be greater than or equal to twice the size needed to house an embedded token, $d_{model} \geq 2 \cdot d_{embed}$. In practice, some embeddings may be losslessly compressible. However, this assumption does not hold for all embeddings.

However, direct RNN simulation is sufficient, but not necessary, for Turing completeness. Therefore, the size requirement for RNN simulation does not imply an equivalent size requirement for Turing completeness. However, the more general $d_{model} > d_{embed}$ does hold. 

To see that this is the case, assume that the base embedding is not compressible. Now assume $x_t$ and a state variable representing the internal state of a Turing machine is compressed into a latent sequence presented to an FFN. Assume the Turing machine's internal state may be compressed into a single binary value as a lower bound. The minimum dimensionality of the latent vector containing the Turing machine state and $x_t$ is $d_{embed} + 1$. Recall, the FFN input dimensionality is required to be identical to $d_{model}$.

Therefore, for a decoder-only transformer model to be Turing complete, it must be true that $d_{model} > d^*_{embed}$ with $d^*_{embed}$ being the dimsensionality of the compressed token embedding.

Interestingly, the inability to recognize PARITY shown in \cite{hahn2020theoretical} may be duplicated by showing that arbitrarily long binary words aren't compressible to any fixed size $d$. Consider, if the input to an attention layer is an $n$ token sequence with each token encoding a binary value, at most $2^d$ values may be losslessly compressed for PARITY computation. Therefore, in the case of hard attention without auto-regression, PARITY is not feed-forward recognizable if the length of the binary sequence is greater than $2^d$. 

\subsection{Transformers and Wang's B Machines}

It is important to point out, decoder-only transformer models do not directly approximate the behavior of Turing machines. Rather, they are computationally much more similar to the B machines studied by \cite{wang1957variant} which have a single tape and are incapable of erase or overwrite. By simulating a Turing machine via B machine, Wang showed that erasure is not necessary for computational universality. However, he also showed that a B machine cannot generate tape content identical to that of a Turing machine in all cases due to the lack of overwrite. 

Decoder-only transformers possess additional limitations beyond those imposed on B machines. While they may read from any past tape location, (1) they are incapable of writing to any position on the tape apart from the next available location and (2) they may not read any position on the tape beyond the current write pointer location. This constitutes an unexplored type of theoretical computational machine which we refer to as causal B machines.

\subsection{Parameter Inefficiency Provenance Conjecture}

Based on the proof herein, small decoder-only transformers are computationally universal. However, due to the significant limitations on causal B machines, format restrictions imposed by an application (like sequence-to-sequence modeling) may prevent the architecture from utilizing arbitrary recursion to perform Turing complete computation. Given a single tape and single permissible write location, intermediate computations which do not fit the application output format will either violate the application or the application output format will prevent the intermediate computation result from being written. 

We conjecture that the strong link between model size and model effectiveness may be related to application induced limitations which force the decoder-only model to induce more sophisticated operations rather than learning to compose them from ``basic steps" unfolded through recursion. This is empirically plausible given the emergence of chain of thought \cite{wei2022chain} as a viable option in the largest of models. Our future work will address this question more thoroughly. 

\section{Conclusion}

We have shown that the decoder-only transformer architecture is capable of simulating an arbitrary RNN and is therefore computationally universal under reasonable assumptions. This result holds even for a 1 layer transformer with a single attention head so long as the model dimensionality exceeds the dimensionality of the minimum token embedding. 

However, this result is limited by the fact that the analysis does not consider the limitations imposed by sequence-to-sequence modeling on the output format which may impact the \textit{in situ} computational expressivity of the architecture.

Therefore, future work seeking to improve the parameter efficiency of decoder-only transformers should consider the effect of output format restrictions and potential architectural changes. Changes, like the inclusion of an additional tape (decoder output location), may permit recursion without dimishing the model's aptitude as a language model.


\bibliographystyle{ieeetr}
\bibliography{main}

\end{document}